\documentclass[a4paper]{article}

\usepackage{times}
\usepackage{graphicx}

\usepackage{hyperref}       % hyperlinks
\usepackage{url}            % simple URL typesetting
\usepackage{booktabs}       % professional-quality tables
\usepackage{amsfonts}       % blackboard math symbols
\usepackage{nicefrac}       % compact symbols for 1/2, etc.
\usepackage{microtype}      % microtypography
\usepackage{amsmath,amssymb}
\usepackage{bbm}
\usepackage{graphicx}
\usepackage{color}
\usepackage{bbm}
\usepackage{multirow}
\usepackage{url}
\usepackage{balance}
\usepackage{multibib}
\usepackage{multicol}
\usepackage{mathtools}
\usepackage{arydshln}
\usepackage{bbm}
\usepackage{balance}
\usepackage{algorithm}
\usepackage[noend]{algpseudocode}

\usepackage{pgfplots}
\usepackage{tikz}
\usepackage{standalone}
\usetikzlibrary{patterns}

\usepackage{natbib}
\newcommand{\p}{{\mathbf{p}}}
\newcommand{\q}{{\mathbf{q}}}
\newcommand{\m}{{\mathbf{M}}}

\newcommand{\concat}{\texttt{Concat}}

\newcommand{\mono}{\texttt{Mono}}

\newcommand{\AggregUL}{\texttt{MV-MV}}

\newcommand{\Aggreg}{\texttt{Fusion}}

\newcommand{\bregboost}{\texttt{M$\omega$MvC}$^2$}
\newcommand{\rboost}{\texttt{rBoost.SH}}
\newcommand{\mvwab}{\texttt{MVWAB}}

\newcommand{\sign}{\operatorname{sign}}
\newcommand{\pwr}{^}

\newtheorem{definition}{Definition}

\usepackage{fancyhdr}
\usepackage{fancybox}
\newcommand{\hypos}{{\cal H}}
\newcommand{\inpts}{{\cal X}}
\newcommand{\obs}{{\bf x}}
\newcommand{\view}{v}

\newcommand{\Trn}{\mathcal{S}}

\newcommand{\obsel}[1]{x^{#1}}
\newcommand{\obsv}{\obsel{\view}}

\newcommand{\Predicat}[1]{\mathbbm{1}_{#1}}
\newcommand{\Loss}{\mathcal{L}}
\newcommand{\ro}{\boldsymbol{\rho}}
\renewcommand{\P}{\boldsymbol{\Pi}}

\renewcommand{\v}{\mathbf{v}}
\newcommand{\pp}{\boldsymbol{\pi}}
\newcommand{\MVBoost}{\texttt{M$\omega$MvC}$^2${}}
\newcommand{\Reuters}{\texttt{Reuters}}
\newcommand{\MINS}[1]{\texttt{MNIST}$_{#1}$}

\newcommand{\nviews}{V}
\newcommand{\X}{\mathcal{X}}
\newcommand{\Y}{\mathcal{Y}}
\newcommand{\R}{\mathbb{R}}
\newcommand{\prods}[2]{\left\langle #1,#2\right\rangle}

\newcommand{\ver}{1}
\title{Multiview Learning  of Weighted Majority Vote by Bregman Divergence Minimization} 
\author{Anil Goyal$^{1,2}$ \and Emilie Morvant$^1$ \and  Massih-Reza Amini$^2$ \and 
	$\mbox{}^1$ \small Univ Lyon, UJM-Saint-Etienne, CNRS, Institut d'Optique Graduate School, \\ \small  Laboratoire Hubert Curien UMR 5516, F-42023, Saint-Etienne, France
	\and
	$\mbox{}^2$ \small Univ. Grenoble Alps, Laboratoire d'Informatique de Grenoble, AMA, \\ \small Centre Equation 4, BP 53, F-38041 Grenoble Cedex 9, France
}

\begin{document}

\maketitle              

\begin{abstract}
  We tackle the issue of classifier combinations when observations have multiple views.
  Our method jointly learns view-specific weighted majority vote classifiers ({\it i.e.} for each view) over a set of base voters, and a second weighted majority vote classifier over the set of these view-specific weighted majority vote classifiers.
  We show that the empirical risk minimization of the final majority vote  given a multiview training set can be cast as the minimization of Bregman divergences.
  This allows us to derive a parallel-update optimization algorithm for learning our multiview model.
  We empirically study our algorithm with a particular focus on the impact of the training set size on the multiview learning results. 
 % on a multilingual extension of the \Reuters{} RCV1/RCV2 corpus using five different languages as well as \MINS{1} and \MINS{2} collections. 
%Our experiments show that the proposed approach is especially effective in the interesting situation where there are few labeled multiview observations available.
%Our experiments show that state-of-the art multiview majority vote based approaches we used are able to overcome the lack of labeled information and that the proposed approach is still more effective in this way.
  The experiments show that our approach is able to overcome the lack of labeled information.

\end{abstract}

\section{Introduction}

In many real-life applications, observations are produced by more than one source and are so-called multiview~\cite{sun2013survey}.  
For example, in  multilingual regions of the world, including many regions of Europe or in Canada,  documents are available in more than one language. 
The aim of multiview learning is to use this multimodal information by combining the predictions of each classifier (or the models themselves) operating over each view (called view-specific classifier) in order to improve the overall performance beyond that of predictors trained on each view separately, or by combining directly the views~\cite{Early-Late-ACMMultimedia05}.

\medskip
\noindent \textbf{Related works.} The main idea here follows the conclusion of the seminal work of \textit{Blum and Mitchell}~\cite{blum98} which states that correlated yet not completely redundant views contain valuable information for learning.
Based on this idea, many studies on multiview learning have been conducted and they can be grouped in three main categories. 
These approaches exploit the redundancy in different representations of data, either by projecting the view-specific representations in a common canonical space~\cite{Gonen11,Zhang11,XuTao14}, or by constraining the classifiers to have \emph{similar} outputs on the same observations; for example by adding a disagreement term in their objective functions~\cite{Sindawani:08}, or lastly by exploiting diversity in the views in order to learn the final classifier defined as the majority vote over the set of view-specific classifiers~\cite{Peng11,Peng17,Xiao12}. 
While the two first families of approaches were designed for learning with labeled and unlabeled training data, the last one,  were developed in the context of supervised learning.
In this line, most of the supervised multiview learning algorithms dealt with the particular case of two view learning~\cite{Farquhar05,janodet:hal-00403242,emvboost}, and some recent works studied the general case of multiview learning with more than two views under the majority vote setting. 
\textit{Amini et al.} \cite{Amini09} derived a generalization error bound for classifiers learned on multiview examples and identified situations where it is more interesting to use all views to learn a uniformly weighted majority vote classifier instead of single view learning. 
\textit{Ko\c co et al.} \cite{Koco11} proposed a Boosting-based strategy that maintains a different distribution of examples with respect to each view. 
For a given view, the corresponding distribution is updated based on view-specific {\it weak} classifiers from that view and all the other views with the idea of using all the view-specific distributions to weight hard examples for the next iteration. 
\textit{Peng et al.} \cite{Peng11,Peng17} enhanced this idea by maintaining a single weight distribution among the multiple views in order to ensure consistency between them. 
\textit{Xiao et al.} \cite{Xiao12} proposed a multiview learning algorithm where they boost the performance of view-specific classifiers by combining multiview learning with Adaboost.

\medskip
\noindent \textbf{Contribution.} In this work, we propose a multiview Boosting-based algorithm, called \bregboost, for the general case where observations are described by more than two views. 
Our algorithm combines previously learned view-specific classifiers as in~\cite{Amini09} but with the difference that it jointly learns two sets of weights for, first, combining view-specific {\it weak classifiers}; and then combining the obtained view-specific weighted majority vote classifiers to get a final weighted majority vote classifier. 
%Note that~\cite{Goyal2017} have recently proven theoretically that this kind of double weighted majority vote is well-suited for tackling multiview learning task, the main difference with our setting being that their weights are restricted to be distributions over the view-specific classifiers.
We show that the minimization of the classification error over a multiview training set can be cast as the minimization of Bregman divergences  allowing the development of an efficient parallel update scheme to learn the weights. 
Using a large publicly available corpus of multilingual documents extracted from the \Reuters{} RCV1 and RCV2 corpora as well as \MINS{1} and \MINS{2} collections, we show that our approach consistently improves over other methods, in the particular when there are only few training examples available for learning. This is a particularly interesting setting when resources are limited, and corresponds, for example, to the common situation of multilingual data.

\medskip
\noindent \textbf{Organization of the paper.} In the next section, we present the double weighted majority vote classifier for multiview learning.  Section \ref{sec:model} shows that the learning problem is equivalent to a Bregman-divergence minimization and describes the Boosting-based algorithm we developed to learn the classifier. In Section \ref{sec:results}, we present experimental results obtained with our approach. %on a subcollection of the \Reuters{} RCV1/RCV2 corpus and \MINS{1} and \MINS{2} collections. 
Finally, in Section \ref{sec:Conc} we discuss the outcomes of this study and give some pointers to further research.

\section{Notations and Setting}

\begin{figure}[t!]
	\centering
	\fbox{%
\begin{tikzpicture}

%
%\begin{scope}[xshift=0.0cm,yshift=0cm]
%\node [draw=black,thick,rectangle,minimum width=15.6cm,minimum height=10.6cm] {};
%\end{scope}

%=========UPPER LEVEL OF HIERARCHY========%
\begin{scope}[yshift=3.6cm]

\node[text width=1cm,black] at (0.2,2.4) {\mbox{\large{all views}}};

\begin{scope}[xshift=1.9cm,yshift=0.0cm]
\node [pattern=north west lines,draw=black,thick,rectangle,minimum width=1.0cm,minimum height=1.6cm] {};
\end{scope}

\begin{scope}[xshift=7.0cm,yshift=1.0cm]
\node [pattern=north west lines,draw=black,thick,rectangle,minimum width=1.0cm,minimum height=3.6cm] {};
\end{scope}

\begin{scope}[xshift=11.5cm,yshift=0.45cm]
\node [pattern=north west lines, draw=black,thick,rectangle,minimum width=1.0cm,minimum height=2.5cm] {};
\end{scope}

%Hyper-posterior and hyper-prior distribution
%\draw[smooth,very thick,color=blue,domain=-1:18.5] plot (\x,{1/(0.1*sqrt(2*pi))*exp(-((\x-9)^2)/(2*3.5^2))}) node[above] {};
%\draw[smooth,very thick,color=red,domain=-1:18.5] plot (\x,{1/(0.07*sqrt(2*pi))*exp(-((\x-7)^2)/(2*2.5^2))}) node[above] {};

%\node[text width=1cm,red] at (4.9,4) {\mbox{\huge$\rho$}};
%\node[text width=1cm,blue] at (6.4,3.2) {\mbox{\huge$\pi$}};

% Plotting axis with dots
%\draw[very thick,-] (-1,0) -- (18.5,0) node[anchor=north west] {};
%\filldraw [black] (1.7,0) circle (2pt);
%\filldraw [black] (7.7,0) circle (2pt);
%\filldraw [black] (15.9,0) circle (2pt);
\end{scope}

%=========BASE LEVEL OF HIERARCHY========%

%%%%% v=1 %%%%%%%%
             %Distributions
%\draw[smooth,very thick,color=blue,domain=-1:4.5] plot (\x,{2*1/exp(((\x-2.1)^2)/2)})  {};
%\draw[smooth,very thick,color=red,domain=-1:4.5] plot (\x,{2.5*1/exp(((\x-1.7)^2)/2)})  {};

\node[text width=3cm,black] at (1.3,2.45) {\mbox{\large$v = 1$}};
%\node[text width=3cm,blue] at (2.6,1) {\mbox{\Large$P_1$}};

\begin{scope}[xshift=0.5cm,yshift=0.8cm]
\node [fill=black,draw=black,thick,rectangle,minimum width=.6cm,minimum height=1.6cm] {};
\end{scope}

\begin{scope}[xshift=1.5cm,yshift=1.12cm]
\node [fill=black,draw=black,thick,rectangle,minimum width=.6cm,minimum height=2.3cm] {};
\end{scope}

\begin{scope}[xshift=2.5cm,yshift=0.5cm]
\node [fill=black,draw=black,thick,rectangle,minimum width=.6cm,minimum height=1.0cm] {};
\end{scope}

\begin{scope}[xshift=3.5cm,yshift=0.9cm]
\node [fill=black,draw=black,thick,rectangle,minimum width=.6cm,minimum height=1.8cm] {};
\end{scope}

			%Axis
%\draw[very thick,-] (-1,0) -- (4.5,0) node[midway,below] {};
%${\mathcal{H}_1 = \{h_1^1,h_2^1, \dots, h_{n_1}^1\}}$
\filldraw [black] (.5,0) circle (1.5pt) node[below] {\mbox{\Large$h_{1,1}$}};
\filldraw [black] (1.5,0) circle (1.5pt) node[below] {\mbox{\Large$h_{1,2}$}};
%\filldraw [black] (1.5,-0.35) circle (0.5pt);
%\filldraw [black] (1.8,-0.35) circle (0.5pt);
%\filldraw [black] (2.1,-0.35) circle (0.5pt);
%\filldraw [black] (2.4,-0.35) circle (0.5pt);
%\filldraw [black] (2.7,-0.35) circle (0.5pt);
%\filldraw [black] (3,-0.35) circle (0.5pt);
\filldraw [black] (2.5,0) circle (1.5pt) node[below] {\mbox{\Large$h_{1,3}$}};
\filldraw [black] (3.5,0) circle (1.5pt) node[below] {\mbox{\Large$h_{1,4}$}};
			%Rectangle
\begin{scope}[xshift=2.0cm,yshift=1cm]
\node [draw,thick,rectangle,minimum width=4.6cm,minimum height=3.6cm] {};
\end{scope}

%%%%%%%  v=2 %%%%%%
             %Distributions
%\draw[smooth,very thick,color=blue,domain=5:10.5] plot (\x,{2*1/exp(((\x-7.2)^2)/2)}) node[right] {};
%\draw[smooth,very thick,color=red,domain=5:10.5] plot (\x,{2.2*1/exp(((\x-7.7)^2)/2)}) node[right] {};

%\node[text width=3cm,blue] at (6.8,1) {\mbox{\Large$P_2$}};

			%Axis

\begin{scope}[xshift=6cm,yshift=0cm]

\begin{scope}[xshift=-0.8cm,yshift=0.4cm]
\node [fill=black,draw=black,thick,rectangle,minimum width=.6cm,minimum height=.8cm] {};
\end{scope}

\begin{scope}[xshift=0.2cm,yshift=0.8cm]
\node [fill=black,draw=black,thick,rectangle,minimum width=.6cm,minimum height=1.6cm] {};
\end{scope}

\begin{scope}[xshift=1.2cm,yshift=1.2cm]
\node [fill=black,draw=black,thick,rectangle,minimum width=.6cm,minimum height=2.4cm] {};
\end{scope}

\begin{scope}[xshift=2.1cm,yshift=1.0cm]
\node [fill=black,draw=black,thick,rectangle,minimum width=.6cm,minimum height=2.0cm] {};
\end{scope}

\begin{scope}[xshift=3.0cm,yshift=0.89cm]
\node [fill=black,draw=black,thick,rectangle,minimum width=.6cm,minimum height=1.8cm] {};
\end{scope}

\node[text width=3cm,black] at (.1,2.45) {\mbox{\large$v = 2$}};
%\draw[very thick,-] (-1,0) -- (4.5,0) node[midway,below] {};
%${\mathcal{H}_1 = \{h_1^1,h_2^1, \dots, h_{n_1}^1\}}$
\filldraw [black] (-0.8,0) circle (1.5pt) node[below] {\mbox{\Large$h_{2,1}$}};
\filldraw [black] (.2,0) circle (1.5pt) node[below] {\mbox{\Large$h_{2,2}$}};
%\filldraw [black] (1.5,-0.35) circle (0.5pt);
%\filldraw [black] (1.8,-0.35) circle (0.5pt);
%\filldraw [black] (2.1,-0.35) circle (0.5pt);
%\filldraw [black] (2.4,-0.35) circle (0.5pt);
%\filldraw [black] (2.7,-0.35) circle (0.5pt);
%\filldraw [black] (3,-0.35) circle (0.5pt);
\filldraw [black] (1.2,0) circle (1.5pt) node[below] {\mbox{\Large$h_{2,3}$}};
\filldraw [black] (2.1,0) circle (1.5pt) node[below] {\mbox{\Large$h_{2,4}$}};
\filldraw [black] (3.0,0) circle (1.5pt) node[below] {\mbox{\Large$h_{2,5}$}};
\end{scope}

			%Rectangle
\begin{scope}[xshift=7.1cm,yshift=1cm]
\node [draw,thick,rectangle,minimum width=5.2cm,minimum height=3.6cm] {};
\end{scope}

%%%%% v=3 %%%%%%
             %Distributions
%\draw[smooth,very thick,color=blue,domain=13.5:18.5] plot (\x,{1.6*1/exp(((\x-16)^2)/2)}) node[right] {};
%\draw[smooth,very thick,color=red,domain=13.5:18.5] plot (\x,{2.2*1/exp(((\x-16.2)^2)/2)}) node[right] {};
%\node[text width=1cm,red] at (18.1,1) {\mbox{\Large$Q_V$}};
%\node[text width=1cm,blue] at (16.7,1) {\mbox{\Large$P_V$}};

			%Axis
\begin{scope}[xshift=10.5cm,yshift=0cm]
\begin{scope}[xshift=0.0cm,yshift=0.89cm]
\node [fill=black,draw=black,thick,rectangle,minimum width=.6cm,minimum height=1.8cm] {};
\end{scope}

\begin{scope}[xshift=1.0cm,yshift=0.5cm]
\node [fill=black,draw=black,thick,rectangle,minimum width=.6cm,minimum height=1.0cm] {};
\end{scope}

\begin{scope}[xshift=2.0cm,yshift=1.2cm]
\node [fill=black,draw=black,thick,rectangle,minimum width=.6cm,minimum height=2.4cm] {};
\end{scope}

\node[text width=3cm,black] at (1.0,2.45) {\mbox{\large$v = 3$}};
%\draw[very thick,-] (-.5,0) -- (4.5,0) node[midway,below] {};
%${\mathcal{H}_1 = \{h_1^1,h_2^1, \dots, h_{n_1}^1\}}$
\filldraw [black] (0,0) circle (1.5pt) node[below] {\mbox{\Large$h_{3,1}$}};
\filldraw [black] (1,0) circle (1.5pt) node[below] {\mbox{\Large$h_{3,2}$}};
%\filldraw [black] (1.5,-0.35) circle (0.5pt);
%\filldraw [black] (1.8,-0.35) circle (0.5pt);
%\filldraw [black] (2.1,-0.35) circle (0.5pt);
%\filldraw [black] (2.4,-0.35) circle (0.5pt);
%\filldraw [black] (2.7,-0.35) circle (0.5pt);
%\filldraw [black] (3,-0.35) circle (0.5pt);
\filldraw [black] (2,0) circle (1.5pt) node[below] {\mbox{\Large$h_{3,3}$}};
\end{scope}

			%Rectangle
\begin{scope}[xshift=11.5cm,yshift=1cm]
\node [draw,thick,rectangle,minimum width=3.2cm,minimum height=3.6cm] {};
\end{scope}

%%Connecting Dots between v=1 and v=V views.
%\filldraw [black] (10.9,1) circle (2pt);
%\filldraw [black] (11.3,1) circle (2pt);
%\filldraw [black] (11.7,1) circle (2pt);
%\filldraw [black] (12.1,1) circle (2pt);
%\filldraw [black] (12.5,1) circle (2pt);
%\filldraw [black] (12.9,1) circle (2pt);

\end{tikzpicture}
}
	\caption{\label{fig:MultiviewHierarchy}Illustration of {\MVBoost} with $\nviews{=}3$.
		For all views $\view\in\{1,2,3\}$, we have a set of view-specific weak classifiers  $(\hypos_v)_{1\le \view\le V}$ that are learned over a multiview training set. The objective is then to learn the weights $\P$ (black histograms) associated  to $(\hypos_v)_{1\le \view\le V}$; and the weights $\ro$ (hatched histograms) associated to weighted majority vote classifiers such that the \mbox{$\ro\P$-weighted} majority vote classifier $B_{\rho\Pi}$ (Equation~\ref{eq: majority vote 1}) will have the smallest possible generalization error.}
	
\end{figure}
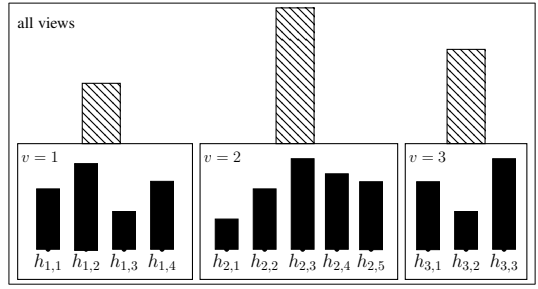

For any positive integer $N$, $[N]$ denotes the set $[N]{\doteq} \{1, \ldots , N\}$. We consider binary classification problems with $\nviews{\geq} 2$ input spaces $\X_v\subset \R^{d_v}; \forall v\in [\nviews]$, and an output space $\Y{=} \{-1,+1\}$.  Each \emph{multiview observation} $\obs\in \X_1{\times}\cdots{\times}\X_{\nviews}$ is a sequence  $\obs {\doteq} (\obsel{1}, \cdots,\obsel{\nviews})$ where each \emph{view} $\obsv$ provides a representation of the same observation in a different vector space $\inpts_v$ (each vector space are not necessarily of the same dimension). We further assume that we have a finite set of {\it weak classifiers} $\hypos_\view{\doteq}\{h_{\view,j}: \inpts_v \rightarrow \{-1,+1\}\mid j\in[n_\view]\}$ of size $n_\view$.
We aim at learning  a two-level encompassed weighted majority vote classifier where at the first level a weighted majority vote is build for each view $\view{\in}[\nviews]$ over the associated set of weak classifiers $\hypos_\view$, and the final classifier, referred to as the Multiview double $\omega$eighted Majority vote Classifier (\MVBoost), is a weighted majority vote over the previous view-specific majority vote classifiers (see Figure~\ref{fig:MultiviewHierarchy} for an illustration).
Given a training set $\Trn{=}(\obs_i,y_i)_{1\le i\le m}$ of size $m$ drawn {\it i.i.d.} with respect to a fixed, yet unknown, distribution ${\cal D}$ over $(\X_1{\times}\cdots{\times}\X_{\nviews}){\times} \Y$, the learning objective is to train the weak view-specific classifiers $(\hypos_\view)_{1\le\view\le \nviews}$ and to choose two sets of weights; $\P=(\pp_\view)_{1\le\view\le \nviews}$, where $\forall \view\in[V],\ \pp_\view{=}(\pi_{\view,j})_{1\le j\le n_v}$, and $\ro{=}(\rho_\view)_{1\le\view\le\nviews}$, such that the $\ro\P$-weighted majority vote classifier $B_{\ro\P}$
\begin{equation}
\label{eq: majority vote 1} B_{\ro\P}(\mathbf{x})=\sum_{\view=1}^\nviews \rho_v\sum_{j=1}^{n_\view} \pi_{\view,j}\, h_{\view,j}(\obsel{\view})
\end{equation}
has the smallest possible generalization error on ${\cal D}$.
We follow the Empirical Risk Minimization principle~\cite{Vapnik99}, and aim at minimizing the \mbox{0/1-loss} over $\Trn$:
\begin{equation*}
\hat \Loss_m^{\mbox{\tiny 0/1}}(B_{\ro\P},\Trn) = 
\frac{1}{m} \sum_{i=1}^{m} \Predicat{ y_i B_{\ro\P}(\mathbf{x}_i) \leq 0 },
\end{equation*}
where $\Predicat{p}$ is equal to $1$ if the predicate $p$ is true, and $0$ otherwise.  
As this loss function is non-continuous and non-differentiable, it is typically replaced by an appropriate convex and differentiable proxy.
Here, we replace $\Predicat{z \leq 0}$ by the logistic upper bound $a\log(1+e^{-z})$, with $a{=}(\log 2)^{-1}$. 
The misclassification cost becomes
\begin{equation}
\label{eq:LogLoss}
\hat \Loss_m(B_{\ro\P},\Trn)  = \frac{a}{m} \sum_{i=1}^{m}  \ln\Big(1 + \exp\big(- y_i B_{\ro\P}(\mathbf{x}_i)\big)\Big),
\end{equation}
and the objective would be then to find the optimal combination weights $\P\pwr\star$ and $\ro\pwr\star$ that minimize this surrogate logistic loss.

\section{An iterative parallel update algorithm to learn \MVBoost}
\label{sec:model}
In this section, we first show how the minimization of the surrogate loss of Equation~\ref{eq:LogLoss} is equivalent to the minimization of a given Bregman divergence. Then, this equivalence  allows us to employ a parallel-update optimization algorithm to learn the weights $\P{=}(\pp_\view)_{1\le\view\le \nviews}$ and $\ro$ leading to this minimization.

\subsection{Bregman-divergence optimization}
\label{sec:BDO}
We first recall the definition of a Bregman divergence \cite{Bregman:67,Lafferty:1999}.
\begin{definition}[Bregman divergence]
\label{def:Bregman-Distance}
Let $\Omega\subseteq \R^m$ and $F : \Omega \rightarrow \mathbb{R}$ be a continuously differentiable and strictly convex real-valued function.
The Bregman divergence $D_F$ associated to $F$ is defined for all $(\p,\q) \in \Omega\times \Omega$ as
\begin{align}
\label{eq:Bregman-Distance}
D_F (\p || \q) \doteq F(\p) - F(\q) - \prods{\nabla F(\q)}{(\p-\q)},
\end{align}
where $\nabla F(\q)$ is the gradient of $F$ estimated at $\q$, and the operator $\prods{\cdot}{\cdot}$ is the dot product function.
\end{definition}
The optimization problem arising from this definition that we are interested in, is to find a vector $\p^\star\in \Omega$---that is the closest to a given vector $\q_0\in \Omega$---under the set $\cal P$ of $V$ linear constraints
\[
\mathcal{P} \doteq\{\p\in \Omega | \forall \view\in[V],\ \rho_v\p^{\top} \m_v=\rho_\view\tilde \p^\top \m_v\},
\] 
where $\tilde \p{\in} \Omega$ is a
specified vector, and $\m_v$ is a $m {\times} n_v$ matrix with $n_v{=}|\hypos_\view|$ the
number of weak classifiers for view $v{\in}[V]$. 
Defining the Legendre transform as
\[
 L_{ F}  \left(  \q, \sum_{v=1}^\nviews  \rho_\view\m_v\pp_\view  \right)  \doteq  \mathop{\arg \min}_{\p \in \Omega} \left\{ D_{ F}(\p||\q)  +  \sum_{\view=1}^\nviews  \prods{\rho_v\m_v\pp_v}{\p}  \right\} .
\]
the dual optimization problem can be stated as finding a vector $\q^\star$ in $\bar{\mathcal{Q}}$, the closure of the set
\begin{equation*}
\mathcal{Q} \doteq\bigg\{\q =  L_{F} \bigg(\q_0, \sum_{v=1}^{V} \rho_v  \m_v \pp_v\bigg) \bigg| \ro\in\R^\nviews; \forall v, \pp_v \in  \mathbb{R} \pwr {n_v}\bigg\},
\end{equation*}
for which $D_F(\tilde \p || \q^\star)$ is the lowest.
It can be shown that both of these optimization problems have the same unique solution~\cite{DellaPietra1997,Lafferty:1999}, with the advantage of having parallel-update optimization algorithms to find the solution  of the dual form in the mono-view case~\cite{Darroch1972,DellaPietra1997,Collins00}, making the use of the latter more appealing.

%\smallskip

According to our multiview  setting and to optimize Equation~\eqref{eq:LogLoss} through a Bregman divergence, we consider the  function $F$ defined for all $ \p{\in} \Omega{=}[0,1]^m$ as
\begin{align*}
F (\p)  \doteq  \sum_{i=1}^{m}  p_i \ln(p_i) + (1- p_i) \ln (1-p_i),
\end{align*}
which from Definition~\ref{def:Bregman-Distance} and the definition of the Legendre transform, yields that for all $(\p, \q)\in \Omega\times \Omega$ and $\mathbf{r}\in\Omega$
\begin{align}
\label{eq:BF1}
&D_F(\p||\q)=\sum_{i=1}^m p_i  \ln\bigg( \frac{p_i}{q_i} \bigg) +(1- p_i)  \ln\bigg( \frac{1-p_i}{1-q_i} \bigg),\\
\label{eq:LF}
&\mbox{and }\forall i\in [m],\ L_F(\q,\mathbf{r})_i = \frac{q_i e^{-r_i}}{ 1 - q_i + q_i e^{-r_i}},
\end{align}
with $a_i$  the $i\pwr{th}$ characteristic of  $\mathbf{a}{=}(a_i)_{1\le i \le m}$ ($\mathbf{a}$ being $\p$, $\q$, $\mathbf{r}$ or $L_F(\q,\mathbf{r})$).
%\smallskip

Now, let  $\q_0=\frac{1}{2}{\textbf 1}_m$ be the vector with all its components set to $\frac{1}{2}$.
For all $i\in[m]$, we define \mbox{$L_F(\q_0,\v)_i = \sigma(v_i)$} with  \mbox{$\sigma(z)=(1+e^{z})^{-1},\ \forall z\!\in \R$}.
We set the matrix $\m_\view$ as for all $(i,j)\in [m]\times[n_\view]$, $(\m_\view)_{ij}=y_i h_{\view,j}(\obsel{\view}_i)$.
Then using  Equations~\eqref{eq:BF1} and~\eqref{eq:LF}, it comes
%Using Equations~\eqref{eq:BF1} and~\eqref{eq:LF}, and setting $\q_0=\frac{1}{2}{\textbf 1}_m$, the vector with all its components set to $\frac{1}{2}$,  making that $\forall i\in[m],\ L_F(\q_0,\v)_i \!=\! \sigma(v_i)$, with $\forall z\!\in \!\R, \sigma(z)=(1+e^{z})^{-1}$, and defining the matrix $\m_\view$ as $\forall i\in [m] \text{ and } j\in [n_\view],\ (\m_\view)_{ij}=y_i h_{\view,j}(\obsel{\view}_i)$; gives~:
\begin{align}
&D_{F} \! \left( \! \mathbf{0}   \Big| \Big|  L_F \left( \!  \q_0,  \sum_{v=1}^{V} \rho_v \m_v \pp_v \! \right) \! \right)
\! = \!\sum_{i=1}^{m} \ln \left( \! 1 \! + \! \exp \!\left(\! -y_i \sum_{v=1}^{V} \rho_v \sum_{j=1}^{n_v} \pi_{v,j} h_{\view,j}(\obsel{\view}_i)\! \right)\! \right).  \label{eq:BF}  %\nonumber
\end{align}
%\smallskip
As a consequence, minimizing Equation~\eqref{eq:LogLoss} is equivalent to minimizing $D_F(\mathbf{0}||\q)$ over $\q\in \bar{\mathcal{Q}_0}$, where for $\Omega=[0,1]^{m}$
\begin{equation}
\label{eq:Q}
   \mathcal{Q}_0 = \left\{ \q\in\Omega \Bigg| q_i = \sigma \left( y_i \sum_{v=1}^{V}  \rho_v  \sum_{j=1}^{n_v} \pi_{v,j} h_{\view,j}(\obsel{\view}_i)  \right)  ; \ro, \P\right\}.
\end{equation}
For a set of weak-classifiers $(\hypos_v)_{1\le \view\le \nviews}$ learned over a training set $\Trn$; this equivalence allows us to adapt the parallel-update optimization algorithm described in \cite{Collins00} to find  the optimal weights $\P$ and $\ro$ defining {\MVBoost} of Equation~\eqref{eq: majority vote 1}, as described in Algorithm \ref{alg:bregboost}.

\subsection{A multiview parallel update algorithm}
\label{sec:Algo}
\begin{algorithm}[t!]
	\caption{Learning \MVBoost}\label{alg:bregboost}
\label{algo1}
	\textbf{Input: } Training set $\Trn = (\mathbf{x}_i, y_i)_{1\le i\le m}$, where $\forall i, \mathbf{x}_i = (\obsel{1}_i, \dots , \obsel{\nviews}_i)$ and $y_i \in \{-1,1\}$; and a maximal number of iterations $T$.
	
%	\medskip
	
		\textbf{Initialization:}  $\ro^{(1)} \leftarrow \frac{1}{\nviews} \mathbf{1}_\nviews$ and $\forall v,  \mathbf{\pp}^{(1)}_v \leftarrow \frac{1}{n_\nviews}\mathbf{1}_{n_\view}$\newline Train the weak classifiers $(\hypos_\view)_{1\le \view\le \nviews}$ over $\Trn$ 
\newline   For $v \in [V]$ set the $m \times n_v$ matrix $\m_v$ such that $\forall i\in [m],\ \forall j\in [n_\view],\ (\m_v)_{ij} = y_i h_{\view,j}(\obsel{v}_i)$ 
%		\vspace{8pt}
			\begin{algorithmic}[1]		
		\For{$t = 1,\ldots , T$}
		\For{$i = 1,\ldots , m$}
%		\vspace{5pt}						
		\State	 $q_{i}^{(t)} = \sigma\left(y_i\displaystyle \sum_{v=1}^{V} \rho_v^{(t)} \sum_{j=1}^{n_v} \pi_{v,j}^{(t)} \  h_{v,j}(\obsel{v}_i)  \right)$
		\EndFor
		
		\For{$v = 1,\ldots , V$}
		\For{$j = 1,\ldots , n_v$}
%		\vspace{5pt}						
		\State $W_{v,j}^{(t)+}= \sum_{i:\sign ((\m_v)_{ij}) = +1 }^{} \  q_{i}^{(t)} | (\m_v)_{ij}|$ 
%		\vspace{2pt}
		\State $W_{v,j}^{(t)-}= \sum_{i:\sign ((\m_v)_{ij}) = -1 }^{} \  q_{i}^{(t)} | (\m_v)_{ij} |$ 
%		\vspace{2pt}
		\State $\delta_{v,j}^{(t)} = \frac{1}{2} \ln \bigg( \frac{W_{v,j}^{(t)+}}{W_{v,j}^{(t)-}} \bigg)$
		
		\EndFor
		\State $\pp_v^{(t+1)} = \pp_v^{(t)} + \boldsymbol{\delta}_{v}^{(t)}$
		
		\EndFor
		\State \textbf{Set} $\ro^{(t+1)}$, as the solution of :
		\vspace{-10pt}
		\begin{align}
		\mathrm{min}_{\ro} & \qquad  - \sum_{v=1}^{V} \rho_\view \sum_{j=1}^{n_v} \bigg( \sqrt{W_{v,j}^{(t)+}} - \sqrt{W_{v,j}^{(t)-}}\bigg) \pwr 2 \label{eq:Opt}\\
		\vspace{-5pt}
		\mbox{s.t. }& \qquad \sum_{v=1}^{V} \rho_\view =1 , \quad \rho_\view \ge 0 \quad \forall v \in [V]\nonumber 
		\end{align}
		\EndFor

	\end{algorithmic}
	\vspace{-10pt}
			\textbf{Return: }  Weights $\ro^{(T)}$ and $\P^{(T)}$.
\end{algorithm}

Once all view-specific {\it weak classifiers} $(\hypos_\view)_{1\le \view\le \nviews}$ have been trained, we start from an initial point $\q^{(1)}\in\mathcal{Q}_0$ (Eq.~\ref{eq:Q}) corresponding to uniform values of weights \mbox{$\ro^{(1)}=\frac{1}{V}\mathbf{1}_\nviews$} and  $\forall \view\in [V],\ \pp^{(1)}_\view=\frac{1}{n_\view}\mathbf{1}_{n_\view}$.
Then, we iteratively update the weights such that at each iteration $t$, using the current parameters $\ro^{(t)}, \P^{(t)}$ and $\q^{(t)}\in\mathcal{Q}_0$, we seek new parameters $\ro^{(t+1)}$ and $\boldsymbol{\delta}_\view^{(t)}$ such that for 
\begin{equation}
\label{eq:qt}
\q^{(t+1)}=L_F(\q_0,\sum_{v=1}^\nviews  \rho^{(t+1)}_\view\m_v(\pp^{(t)}_\view+~\boldsymbol{\delta}_\view^{(t)})),
\end{equation}
 we get $D_F(0 || \q^{(t+1)})\le D_F(0 || \q^{(t)})$.

%\medskip

Following \cite[Theorem~3]{Collins00}, it is straightforward to show that in this case, the following inequality holds:
\begin{align}
\label{eq:DF}
&D_F(\mathbf{0}||\q^{(t+1)})-D_F(\mathbf{0}||\q^{(t)})\le A^{(t)}\,,\\
\nonumber \mbox{where}\quad &
A^{(t)} = -  \sum_{v=1}^{V}  \rho^{(t+1)}_v  \sum_{j=1}^{n_v}  \bigg(   W_{v,j}^{(t)+}(e^{-\delta_{v,j}^{(t)}}-1) - W_{v,j}^{(t)-}(e^{\delta_{v,j}^{(t)}}-1)  \bigg)^{  2},
\end{align}
with $\forall j\in[n_\view]; W_{v,j}^{(t)\pm}=\sum_{i:\text{sign}((\m_v)_{ij})=\pm 1} q_i^{(t)}|(\m_v)_{ij}|$.

By fixing the set of parameters $\ro^{(t+1)}$;  the parameters $\boldsymbol{\delta}_\view^{(t)}$ that minimize $A^{(t)}$ are defined as $\forall \view\in [V], \forall j\in [n_\view]; \delta^{(t)}_{\view,j}=\frac{1}{2}\ln\left(\frac{W_{v,j}^{(t)+}}{W_{v,j}^{(t)-}}\right)$. 
Plugging back these values into the above equation gives
\begin{equation}
\label{eq:At}
A^{(t)}=- \sum_{v=1}^{V}  \rho^{(t+1)}_v \sum_{j=1}^{n_v} \bigg(\sqrt{W_{v,j}^{(t)+}} - \sqrt{W_{v,j}^{(t)-}}\bigg)^{2}.
\end{equation}
Now by fixing the set of parameters $(W^{(t)\pm}_{v,j})_{v,j}$, the weights $\ro^{(t+1)}$ are found by minimizing Equation~\eqref{eq:At} under the linear constraints $\forall v\in [V], \rho_v \ge 0 \text{ and } \sum_{v=1}^{V} \rho_v = 1$. 
%The pseudo-code of the whole iterative process is shown in Algorithm~\ref{algo1}. 
This alternating optimization of $A^{(t)}$ bears similarity with the block-coordinate descent technique~\cite{Bertsekas99}, where at each iteration, variables are split into two subsets---the set of the active variables, and the set of the inactive ones---and the objective function is minimized along active dimensions while inactive variables are fixed at current values. 

\medskip
 \noindent\textbf{Convergence of Algorithm. }The sequences  of weights $(\P^{(t)})_{t\in\mathbb{N}}$ and $(\ro^{(t)})_{t\in\mathbb{N}}$ found by Algorithm~\ref{algo1} converge to the minimizers of the multiview classification loss (Equation~\ref{eq:LogLoss}), as with the resulting sequence $(\q^{(t)})_{t\in\mathbb{N}}$ (Equation~\ref{eq:qt}),  the sequence $(D_F(\mathbf{0}||\q^{(t)}))_{t\in\mathbb{N}}$ is decreasing and since it is lower-bounded (Equation~\ref{eq:BF}), it converges to the minimum of Equation~\eqref{eq:LogLoss}.

\section{Experimental Results}
\label{sec:results}
We present below the results of the experiments we have performed to evaluate the efficiency of Algorithm~\ref{algo1} to learn the set of weights $\P$ and $\ro$ involved in the definition of the $\ro\P$-weighted majority vote classifier $B_{\ro\P}$ (Equation~\eqref{eq: majority vote 1}).

\subsection{Datasets}

\label{sec:datasets}
\textbf{\MINS{}} is a publicly available dataset consisting of $70,000$ images of handwritten digits distributed over $10$ classes~\cite{Lecun98}. 
%The size of different classes in the number of images is given in Table~\ref{tab:MNIST_corpus}. 
For our experiments, we created 2 multiview collections from the initial dataset. 
Following~\cite{ChenD17}, the first dataset ($\mathtt{MNIST}_1$) was created by extracting $4$ no-overlapping quarters of each image considered as its $4$ views.
The second dataset ($\mathtt{MNIST}_2$) was made by extracting $4$ overlapping quarters from each image as its $4$ views.
We randomly splitted each collection by keeping $10,000$ images for testing and the remaining images for training. 

%\begin{table}[h!]
%\centering
%\begin{tabular}{c|c|c|c|c|c}
%	\hline
%Class & \texttt{zero} &\texttt{one} & \texttt{two} & \texttt{three} & \texttt{four} \\\hline 
%\# Images & 6903 & 7877  &6990  &7141 & 6824 \\\hline 
%\multicolumn{6}{c}{} \\
%\hline
%Class & \texttt{five} & \texttt{six} & 	\texttt{seven} & \texttt{eight} & \texttt{nine} \\\hline
%\# Images  & 6313 & 6876 & 7293 & 6825 & 6958 \\ \hline
%\end{tabular}	
%		\caption{Number of images per class in \MINS{}.}
%		\label{tab:MNIST_corpus}
%\end{table}

% and remaining as training samples. 

%Examples of each dataset is shown in Fig  $1$ and $2$ respectively. We reserve $10,000$ of images as test samples and remaining as training samples. 
%\begin{figure}[t]
%	\label{fig:mnist1}\centering
%	\includegraphics[scale=0.2]{images/eg_Dataset_1.png}
%	\caption{Example of $\mathtt{MNIST}_1$ dataset.}
%\end{figure}
%\begin{figure}[t]
%	\label{fig:mnist2}\centering
%	\includegraphics[scale=0.2]{images/eg_Dataset_2.png}
%	\caption{Example of $\mathtt{MNIST}_2$ dataset.}
%\end{figure}

\medskip

\noindent\textbf{\Reuters{} RCV1/RCV2} is a multilingual text classification data extracted from Reuters RCV1 and RCV2  corpus\footnote{\scriptsize\url{https://archive.ics.uci.edu/ml/datasets/Reuters+RCV1+RCV2+Multilingual,+Multiview+Text+Categorization+Test+collection}}.
It consists of more than $110,000$ documents written in five different languages (English, French, German, Italian and Spanish) distributed over six classes. 
In this paper we consider each language as a view.
%We see different languages as different views of the data. 
%The statistics of this dataset are presented in Table~\ref{tab:Reuters_corpus}.
We reserved $30 \%$ of documents for testing and the remaining for training. % and  remaining as training data. 

%\begin{table}[h!]
%\begin{center}
%\begin{tabular}{l|c c l| c}
%\cline{1-2}\cline{4-5}
%Language & \# Docs & $~~~~$ & Class & \# Docs \\ \cline{1-2}\cline{4-5}
%\texttt{English}    &  18,758   & $~~~~$ &  \texttt{C15}  & 18,816 \\
%\texttt{French}    &  26,648   & $~~~~$ &  \texttt{CCAT} & 21,426 \\
%\texttt{German}    &  29,953   & $~~~~$ &  \texttt{E21} & 13,701 \\
%\texttt{Italian}    &  24,039   & $~~~~$ &  \texttt{ECAT} & 19,198 \\
%\texttt{Spanish}  &  12,342   & $~~~~$ &  \texttt{GCAT} & 19,178 \\ \cline{1-2}
%Total  & 111,740   & $~~~~$ &  \texttt{M11}  & 19,421 \\
%\cline{1-2}\cline{4-5}
%\end{tabular}
%\end{center}
%\caption{Number of documents per language (left) and per class (right) in \Reuters{} RCV1/RCV2 corpus.}
%\label{tab:Reuters_corpus}
%\end{table}

\begin{table}[t]
	\centering
		\caption{Test classification accuracy and \mbox{$F_1$-score} of different approaches averaged over all the classes and over $20$ random sets of $m=100$ labeled examples per training set. Along each column, the best result is in bold, and second one in italic. $^{\downarrow}$ indicates that a result is statistically significantly worse than the best result, according to a Wilcoxon rank sum test with $p < 0.02$.}
	\resizebox{\textwidth}{!}{
		\begin{tabular}{ c| c c  c c c c c  c} 
			\hline
			\multirow{2}{*}{Strategy} &
			\multicolumn{2}{c}{\MINS{1}} &&
			\multicolumn{2}{c}{\MINS{2}} &&
			\multicolumn{2}{c}{\Reuters} \\
			\cline{2-3}\cline{5-6}\cline{8-9}  
			&  Accuracy & $F_1$  && Accuracy & $F_1$ && Accuracy & $F_1$  \\\hline
$\mono$  & $.7827 \pm .008^{\downarrow}$ &  $.4355 \pm .009^{\downarrow}$  && $.7896 \pm .008^{\downarrow}$ &  $.4535 \pm .011^{\downarrow}$ && $.7089 \pm .017^{\downarrow}$ &  $.4439 \pm .007^{\downarrow}$\\
$\concat$  & $.7988 \pm .011^{\downarrow}$ &  $.4618 \pm .015^{\downarrow}$  && $.7982 \pm .017^{\downarrow}$ &  $.4653 \pm .021^{\downarrow}$ && $.6918 \pm .029^{\downarrow}$ &  $.4378 \pm .015^{\downarrow}$\\

$\Aggreg$  & $.8167 \pm .017^{\downarrow}$ &  $.4769 \pm .018^{\downarrow}$  && $.8244 \pm .019^{\downarrow}$ & $.4955 \pm .027^{\downarrow}$ && $.7086 \pm .029^{\downarrow}$ &  $.4200 \pm .021^{\downarrow}$\\

$\texttt{MVMLsp}$  & $.7221 \pm .021^{\downarrow}$ &  $.3646 \pm .019^{\downarrow}$  && $.7669 \pm .032^{\downarrow}$ & $.4318 \pm .025^{\downarrow}$ && $.6037 \pm .020^{\downarrow}$ &  $.3181 \pm .022^{\downarrow}$\\

$\AggregUL$  &  ${.8381} \pm .009^{\downarrow}$ &  ${.5238} \pm .015^{\downarrow}$  && $\textit{.8380} \pm .010^{\downarrow}$ &  ${.5307} \pm .016^{\downarrow}$ && $.7453 \pm .023^{\downarrow}$ &  $.4979 \pm .012^{\downarrow}$\\
$\mvwab$ & $\textit{.8470} \pm .015^{\downarrow} $ &  $\textit{.5704} \pm .012^{\downarrow}$  && ${.8331} \pm .016^{\downarrow}$ &  $\textit{.5320} \pm .011^{\downarrow}$ && $.7484 \pm .017^{\downarrow}$ &  $.5034 \pm .016 ^{\downarrow}$\\
$\rboost$ & $.7580 \pm .011^{\downarrow}$ &  $.4067 \pm .009^{\downarrow}$  && $.8247 \pm .009^{\downarrow}$ &  $.5148 \pm .015^{\downarrow}$ && $\textit{.7641} \pm .014 $ &  $\textit{.5093} \pm .010^{\downarrow}$\\
\bregboost   & $\textbf{.8659} \pm .011$ &  $\textbf{.5914} \pm .015$  && $\textbf{.8474} \pm .012$ &  $\textbf{.5523} \pm .018$ && $\textbf{.7662} \pm .010$ &  $\textbf{.5244} \pm .012$ \\ \hline
		\end{tabular}
	}

		\label{tab:results}
	\end{table}
	
\subsection{Experimental Protocol} % and Results}
In  our experiments, we set up binary classification tasks by using all multiview observations from one class as positive examples and all the others as negative examples.
We reduced the imbalance between positive and negative examples by subsampling the latter in the training sets, and used decision trees as view specific weak classifiers.
We compare our approach to the following seven algorithms. 
\begin{itemize}

\item $\mono$  is the best performing %% mono-view
decision tree model operating on a single view.
%a mono-view decision tree models operating on each view separately.
%The reported results are the one obtained with the best performing mono-view classifier.

	\item $\concat$  is an early fusion approach, %% \cite{Early-Late-ACMMultimedia05}
          where a mono-view decision tree  operates over the concatenation of all views of multiview observations. % by concatenating features from all the views. 
          
          \item $\Aggreg$  is a late fusion  approach, sometimes referred to as stacking,  where view-specific classifiers are trained independently over different views using $60 \%$ of the training examples.
          A final multiview model is then trained over the predictions of the view-specific classifiers using the rest of the training examples.
          
          \item $\texttt{MVMLsp}$~\cite{huusari18} is a multiview metric learning approach, where multiview kernels are learned to capture the view-specific information and relation between the views.
          We kept the  experimental setup of \cite{huusari18} with Nystr{\"o}m parameter $0.24$.\footnote{We used the Python code available from  \scriptsize\url{https://lives.lif.univ-mrs.fr/?page_id=12}}

	  \item $\AggregUL$~\cite{Amini09} is a multiview algorithm where view-specific classifiers are trained over the views using all the training examples. The final model is the uniformly weighted majority vote.
        
	\item $\mvwab$~\cite{Xiao12} is a Multiview Weighted Voting AdaBoost algorithm, where multiview learning and ababoost techniques are combined to learn a  weighted majority vote over view-specific classifiers but without any notion of learning weights over views. 
%	In our experiments, we used decision tree classifier to learn view-specific voters at each iteration and we fixed the maximum number of iterations to $T=100$.

	\item $\rboost$~~\cite{Peng11,Peng17} is a multiview boosting approach where a single distribution over different views of training examples is maintained and, the distribution over the views are updated using the multiarmed bandit framework. %~\cite{Auer03}.
%	At each iteration, the algorithm selects a view according to the current distribution and learns the corresponding view-specific classifier. 
%	In our experiments, we used decision tree classifier to learn view-specific voters at each iteration. 
	For the tuning of parameters, we followed the  experimental setup of~\cite{Peng17}. % and fixed the maximum number of iterations to $T=100$.
        \end{itemize}
        
\Aggreg, \AggregUL{}, \mvwab, and \rboost{} make decision based on some majority vote strategies, as the proposed \bregboost{} classifier. The difference relies on how the view-specific classifiers are combined. 
For \mvwab \  and \rboost{}, we used decision tree model to learn view-specific weak classifiers at each iteration of algorithm and fixed the maximum number of iterations to $T=100$. 
To learn \bregboost{}, we generated the matrix $\m_v$ by considering a set of weak decision tree classifiers with different depths (from $1$ to $\max_{d} - 2 $, where $\max_d $ is maximum possible depth of a decision tree). We tuned the maximum number of iterations by cross-validation which came out to be $T=2$ in most of the cases and that we fixed throughout all of the experiments.
To solve the optimization problem for finding the weights $\ro$ (Equation \ref{eq:Opt}),  we used the Sequential Least SQuares Programming (SLSQP) implementation of scikit-learn~\cite{scikit-learn}, that we also used to learn the  decision trees.
 Results are computed over the test set using the accuracy and the standard \mbox{$F_1$-score}~\cite{Powers:11}, which is the harmonic average of precision and recall. Experiments are repeated $20$ times by each time splitting the training and the test sets at random over the initial datasets. 

\subsection{Results}
Table~\ref{tab:results} reports the results obtained for $m{=}100$ training examples by different methods averaged over all classes and  the $20$ test results obtained over $20$ random experiments\footnote{We also did experiments for \mono, \concat, \Aggreg, \AggregUL{} using Adaboost. The performance of Adaboost for these baselines is similar to that of decision trees.}.
%We use bold face to indicate the highest performance rates, and the symbol $^{\downarrow}$ indicates that performance is significantly worse than the best result, according to a Wilcoxon rank sum test used at a p-value threshold of $0.02$ \cite{StatsRankMethods}.
From these results it becomes clear that late fusion and other multiview approaches (except $\texttt{MVMLsp}$) provide consistent improvements over training independent mono-view classifiers and with early fusion, when the size of the training set is small.
Furthermore, \bregboost{} outperforms the other approaches and compared to the second best strategy the gain in accuracy ({\it resp.} $F_1$-score) varies between $0.2\%$ and $2.2\%$ ({\it resp.} $2.2\%$ and $3.8\%$) across the collections.
These results provide evidence that majority voting for multiview learning is an effective way to overcome the lack of labeled information and that all the views do not have the same strength (or do not bring information in the same way) as the learning of weights, as it is done in \bregboost{}, is much more effective than the uniform combination of view-specific classifiers as it is done in \AggregUL.

\begin{figure*}[t!]%
	\centering
	\begin{tabular}{cc}
		\hspace{-9mm}\includegraphics[scale=.185]{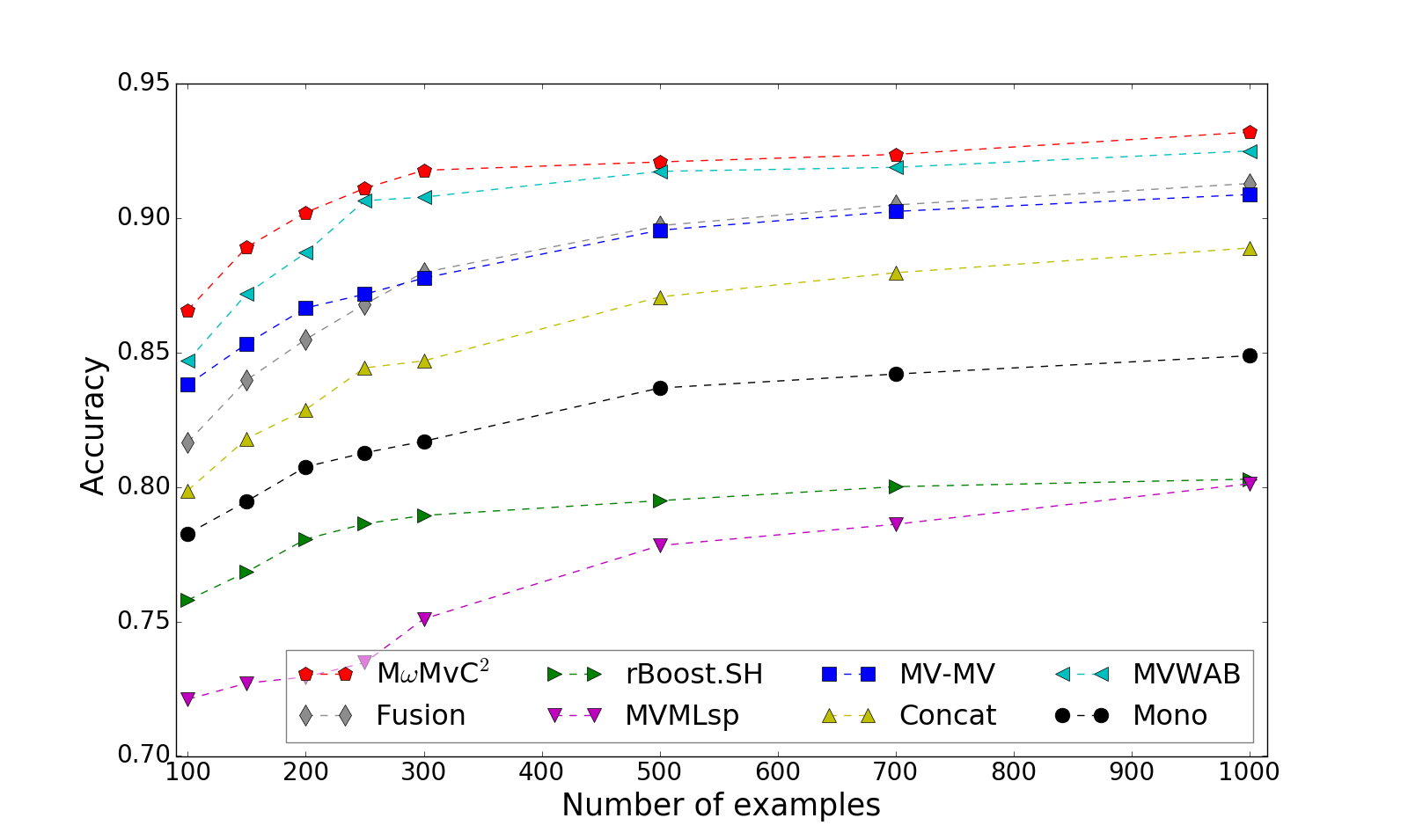}  & 
		\hspace{-0.48cm}\includegraphics[scale=.185]{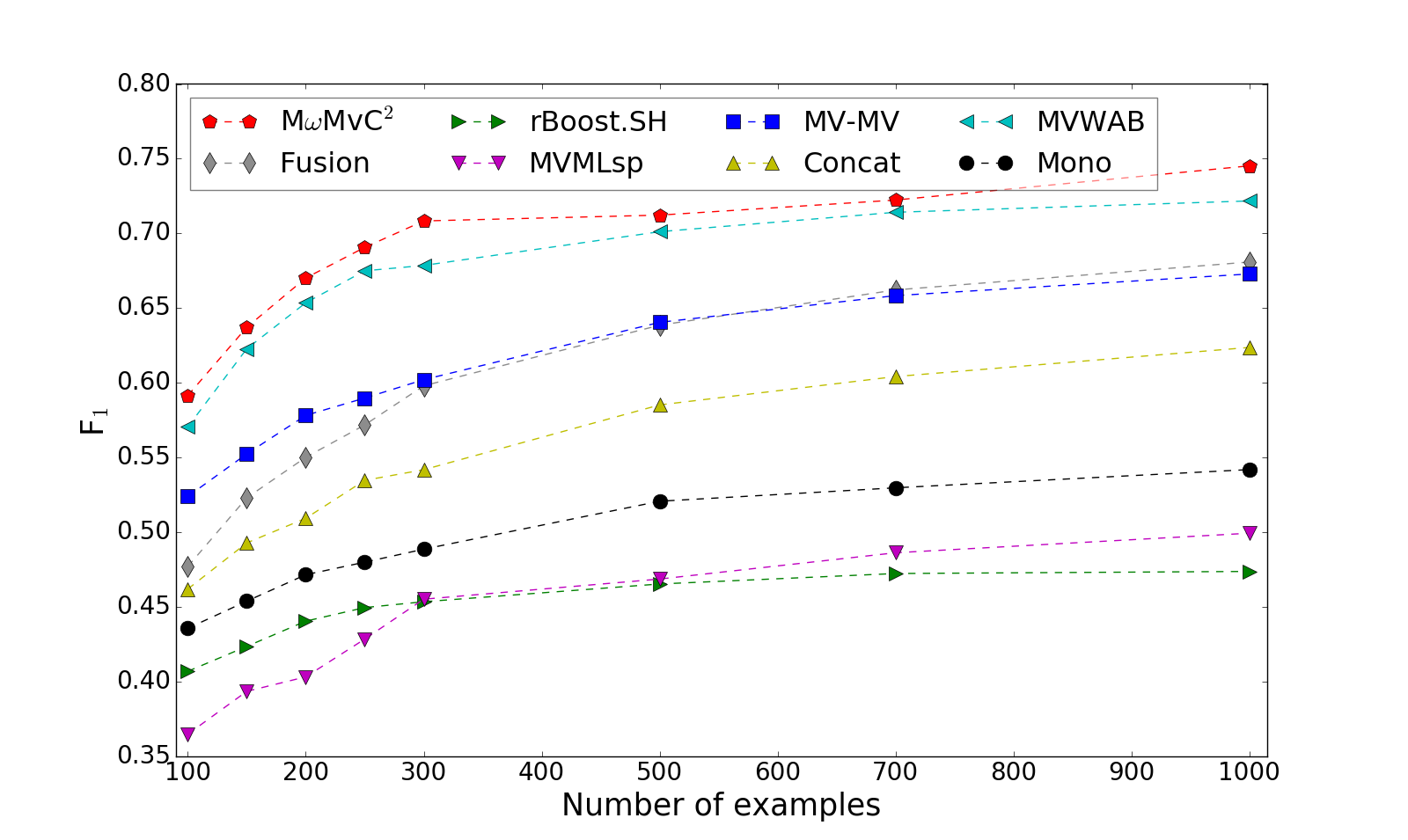}\\
		\multicolumn{2}{c}{(a) \MINS{1}}\\
		\hspace{-9mm}\includegraphics[scale=.185]{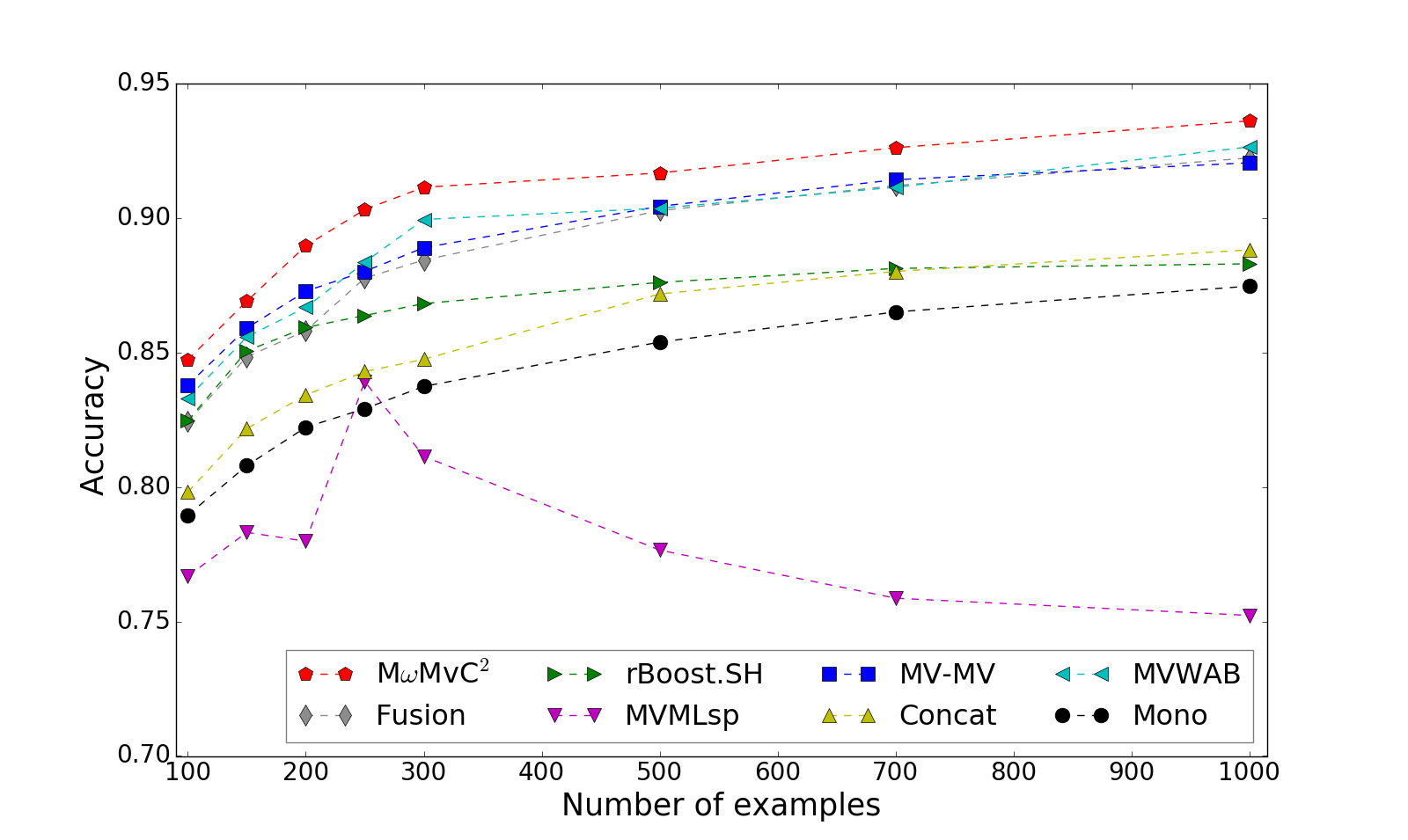}  & 
		\hspace{-0.48cm}\includegraphics[scale=.185]{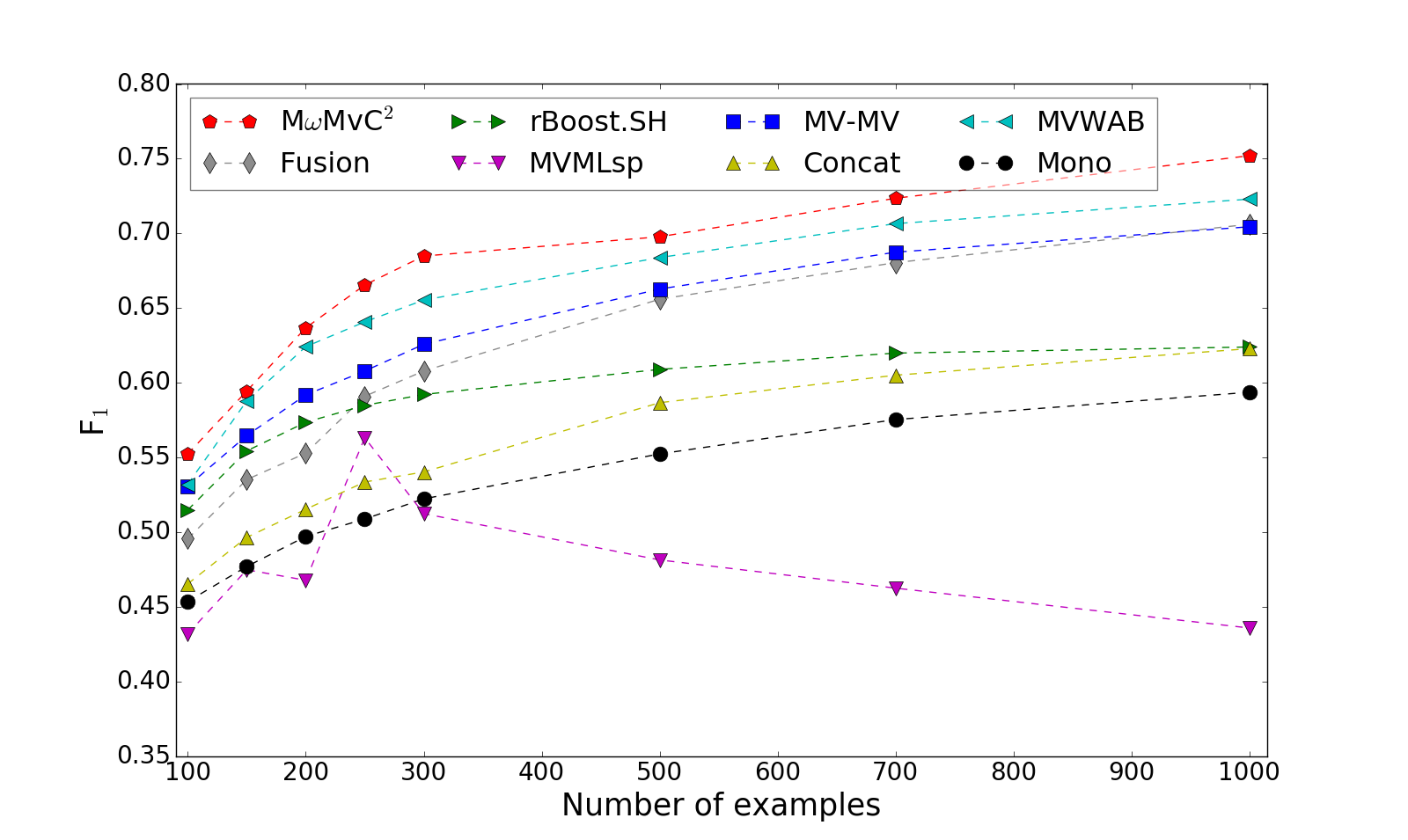} \\
		\multicolumn{2}{c}{(b) \MINS{2}}\\
		\hspace{-9mm}\includegraphics[scale=.185]{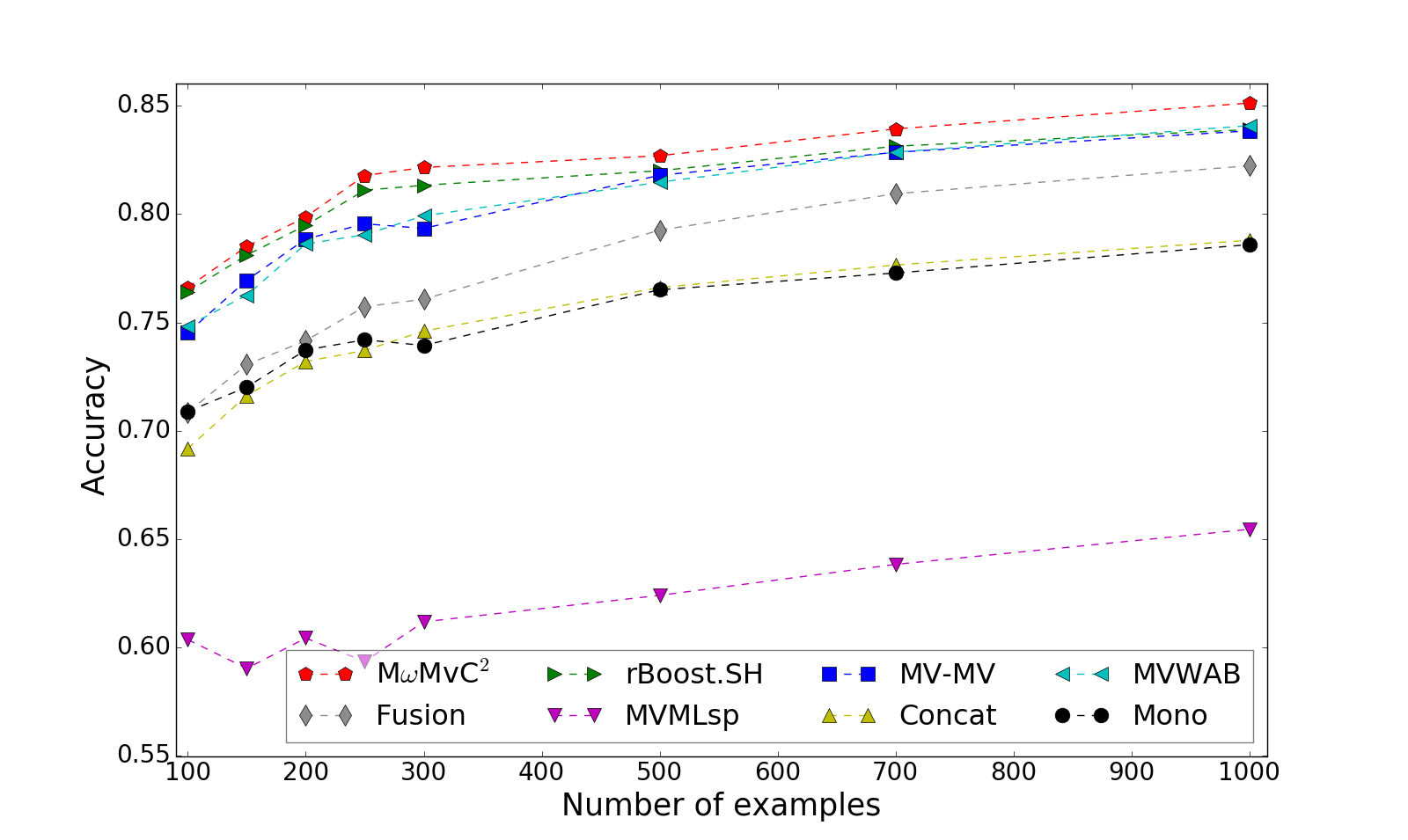}  & 
		\hspace{-0.48cm}\includegraphics[scale=.185]{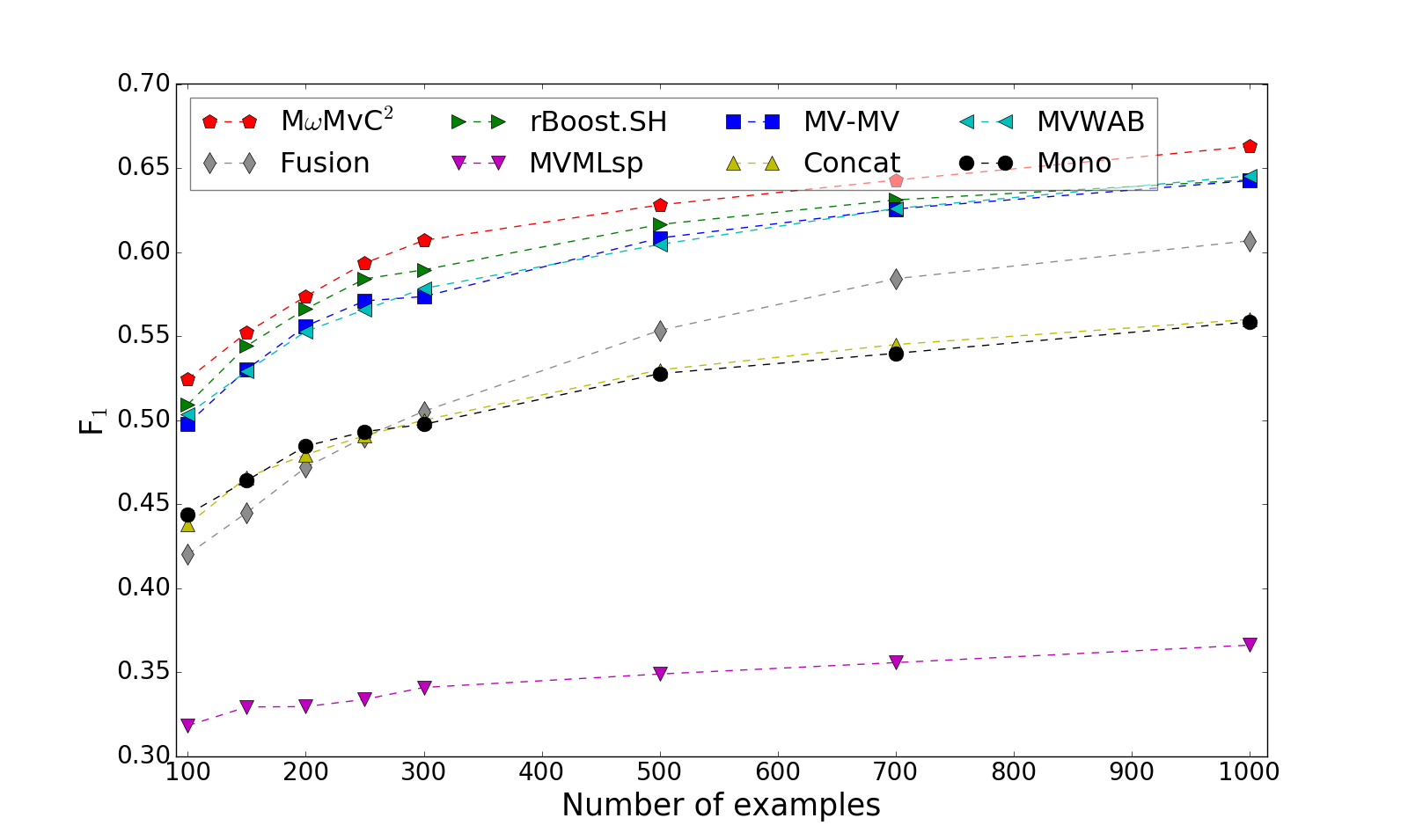} \\
		\multicolumn{2}{c}{(c) \Reuters}\\
	\end{tabular}
	\caption{Evolution of accuracy and $F_1$-score {\it w.r.t} to the number of labeled examples in the initial labeled training sets on \MINS{1}, \MINS{2} and \Reuters{} datasets.}
	\label{fig:plots}
\end{figure*}

We also analyze the behavior of the algorithms for growing initial amounts of labeled data. % in the training set.
Figure~\ref{fig:plots} illustrates this by showing the evolution of the accuracy and the $F_1$-score  with respect to the number of labeled examples in the initial labeled training sets on \MINS{1}, \MINS{2} and \Reuters{} datasets. As expected, all performance curves increase monotonically {\it w.r.t} the additional labeled data. When there are sufficient labeled examples, the performance increase of all algorithms actually begins to slow, suggesting that the labeled data carries sufficient information and that the different views do not bring additional information.

An important point here is that $\rboost$---which takes into account both view-consistency and diversity between views---provides the worst results on \MINS{1} where there is no overlapping between the views, while the weighted majority vote as it is performed in \bregboost{} still provides an efficient model.
Furthermore, $\texttt{MVMLsp}$---which learns multiview kernels to capture views-specific informations and relation between views---performs worst on all the datasets.
We believe that the superior performance of our method stands in our two-level framework. % we follow.
Indeed, thanks to this trick, we are able to consider the view-specific information by learning weights over view-specific classifiers, and  to capture the importance of each view in the final ensemble by learning weights over the views.

%These results suggest that \bregboost{} is able to take advantage of different views and the relation which exists among them even in the case where they are not redundant. 
% Moreover, on image dataset $\texttt{MNIST}$, algorithm $\rboost$ (which takes into account both view-consistency and diversity between views) performs worse than multiview majority vote $\AggregUL$ (Figures\ref{fig:MNIST1} and \ref{fig:MNIST2}).
%On text classification tasks (Figure \ref{fig:Reuters}),  as we increase number of training examples $\rboost$ tends to perform similar to $\AggregUL$.
%It means that by following our strategy of learning double weighted majority vote, we can exploit the additional information provided by multiple views of data even for sufficient number of training examples.
%This shows the potential of our algorithm $\bregboost$ where we considered double weighted majority vote over the set of view-specific classifiers.
%By learning the view-specific weighted majority votes, we are taking into account view-specific informations and by learning the weights over views which leads to double weighted majority vote, we are taking into account relation between different views. 

\subsection{A note on the Complexity of Algorithm}
For each view $v$, the complexity of learning decision tree classifiers is $O(d_v\,m log(m))$. 
We learn the weights over the views by optimizing Equation~\eqref{eq:At} (Step $10$ of our algorithm) using SLSQP method which has time complexity of $O(V^3)$. 
Therefore, the overall complexity is $O( V\,d_v\,m.log(m) + T\,(V^3 + \sum_{v=1}^V m\,n_v) )$. 
Note that it is easy to parallelize our algorithm: by using $V$ different machines, we can learn the view-specific classifiers  and weights over them (Steps $4$ to $9$).

\section{Conclusion}
\label{sec:Conc}
In this paper, we tackle the issue of classifier combination when observations have different representations (or have multiple views).
Our approach jointly learns weighted majority vote view-specific classifiers ({\it i.e.} at the view level) over a set of base classifiers, and a second weighted majority vote classifier over the previous set of view specific weighted majority vote classifiers.
We show that the minimization of the multiview classification error is equivalent to the minimization of Bregman divergences.
This embedding allowed to derive a parallel-update optimization boosting-like algorithm to learn the weights of the double weighted multiview majority vote classifier.
Our results show clearly that our method allows to reach high performance in terms of accuracy and $F_1$-score on three datasets in the situation where few initial labeled training documents are available.
It also comes out that compared to the uniform combination of view-specific classifiers, the learning of weights allows to better capture the strengths of different views.

As future work, we would like to extend our algorithm to the \textit{semi-supervised} case, where one has access to an additionally unlabeled set during the training.
One possible way is to learn a view-specific classifier using pseudo-labels (for unlabeled data) generated from the classifiers trained from other views, {\it e.g.}~\cite{Xu16}.
Moreover, the question of extending our work to the case where all the views are not necessarily available or not complete ({\emph{missing views} or \emph{incomplete views}, {\it e.g.}~\cite{Amini09,xu2015multi}), is very exciting.
One solution could be to adapt the definition of the matrix $\m_v$ to allow to deal with incomplete data; this may be done by considering a notion of diversity to complete $\m_v$. 
%Moreover, we would like to explore  \textit{domain adaptation}  where training and test data are drawn from different distributions.
%An interesting direction would be to bind the data distribution to the different views of the data, as in some recent zero-shot learning approaches~\cite{Socher13}.

\subsection*{Acknowledgments.} This work is partially funded by the French ANR project LIVES ANR-15-CE23-0026-03 and the ``R\'egion  Rh\^{o}ne-Alpes''.

\bibliography{biblio}

\end{document}